\documentclass{article}

\usepackage{microtype}
\usepackage{graphicx}
\usepackage{subcaption}
\usepackage{booktabs} 
\usepackage{tcolorbox}
\usepackage{xcolor}
\usepackage{multirow} 
\usepackage{adjustbox}
\usepackage{tabularx}
\usepackage{booktabs}

\usepackage{hyperref}

\usepackage[preprint]{icml2026}

\usepackage{amsmath}
\usepackage{amssymb}
\usepackage{mathtools}
\usepackage{amsthm}
\usepackage{amsfonts} 

\usepackage[capitalize,noabbrev]{cleveref}

\theoremstyle{plain}

\theoremstyle{definition}

\theoremstyle{remark}

\newcommand{\eg}{\textit{e.g.,}\ }
\newcommand{\ie}{\textit{i.e.}\ }

\newcommand{\fireicon}{\raisebox{-0.2em}{\includegraphics[height=1em]{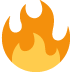}}}
\newcommand{\snowicon}{\raisebox{-0.2em}{\includegraphics[height=1em]{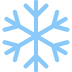}}}

\definecolor{mvscLine}{HTML}{552583}   
\definecolor{mvscFill}{HTML}{E8E2F2}   

\usepackage[textsize=tiny]{todonotes}

\icmltitlerunning{Multimodal Visual Surrogate Compression for Alzheimer’s Disease Classification}

\begin{document}

\twocolumn[
  \icmltitle{Multimodal Visual Surrogate Compression for Alzheimer’s Disease Classification
  }



  \icmlsetsymbol{equal}{*}

  \begin{icmlauthorlist}
    \icmlauthor{Dexuan Ding}{1}
    \icmlauthor{Ciyuan Peng}{2}
    \icmlauthor{Endrowednes Kuantama}{1}
    \icmlauthor{Jingcai Guo}{3}
    \icmlauthor{Jia Wu}{1}
    \icmlauthor{Jian Yang}{1}
    \icmlauthor{Amin Beheshti}{1}
    \icmlauthor{Ming-Hsuan Yang}{4}
    \icmlauthor{Yuankai Qi}{1}
  \end{icmlauthorlist}

  \icmlaffiliation{1}{Macquarie University}
  \icmlaffiliation{2}{Federation University Australia}
  \icmlaffiliation{3}{The Hong Kong Polytechnic University}
  \icmlaffiliation{4}{University of California at Merced}

   \icmlcorrespondingauthor{Yuankai Qi}{yuankai.qi@mq.edu.au}

  \icmlkeywords{Machine Learning, ICML}

  \vskip 0.3in
]



\printAffiliationsAndNotice{}  

\begin{abstract}
High-dimensional structural MRI (sMRI) images are widely used for Alzheimer’s Disease (AD) diagnosis. 
Most existing methods for sMRI representation learning rely on 3D architectures (\eg 3D CNNs), slice-wise feature extraction with late aggregation, or apply training-free feature extractions using 2D foundation models (\eg DINO). 
However, these three paradigms suffer from high computational cost, 
loss of cross-slice relations, and limited ability to extract discriminative features, respectively. 
To address these challenges, we propose Multimodal Visual Surrogate Compression (MVSC). It learns to compress and adapt large 3D sMRI volumes into compact 2D features, termed as visual surrogates,
which are better aligned with frozen 2D foundation models
to extract powerful representations for final AD classification.
MVSC has two key components: a Volume Context Encoder that captures global cross-slice context under textual guidance, and an Adaptive Slice Fusion module that aggregates slice-level information in a text-enhanced, patch-wise manner. 
Extensive experiments on three large-scale Alzheimer’s disease benchmarks demonstrate our MVSC performs favourably on both binary and multi-class classification tasks compared against  state-of-the-art methods.
\end{abstract}

\section{Introduction}

\begin{figure}[h]
    \centering
    \includegraphics[
        trim=10cm 23cm 20cm 12cm,
        clip,
        width=2.0\linewidth
    ]{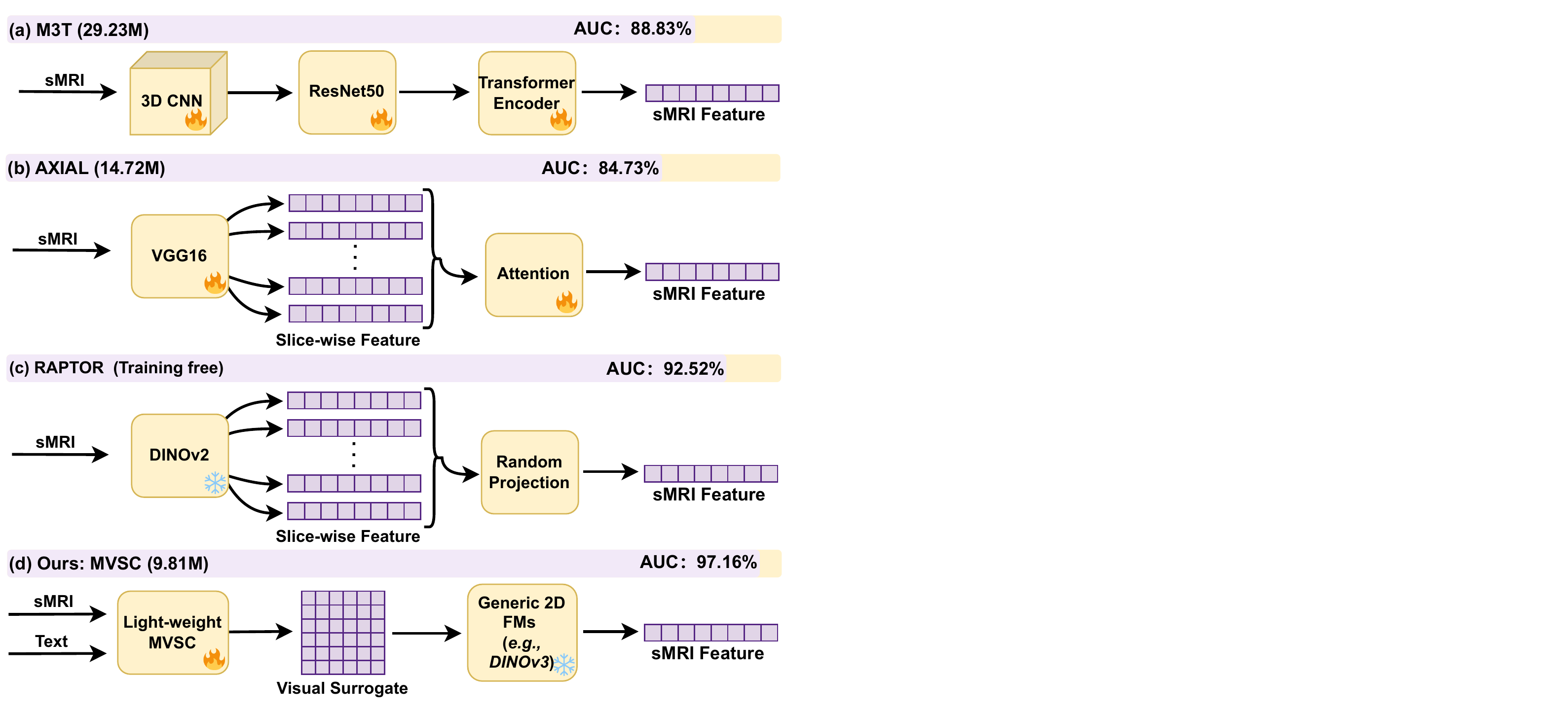}
    \caption{
    Main paradigms of representative sMRI feature extraction for Alzheimer’s disease classification.
    The fire icon \fireicon\ indicates trainable modules and the snowflake icon \snowicon\ indicates frozen modules. 
    (a) Fully volumetric 3D models that directly process sMRI volumes (\eg M3T~\citep{DBLP:conf/cvpr/JangH22}).
    (b) Slice-wise methods with trainable aggregation (\eg AXIAL~\citep{DBLP:journals/corr/abs-2407-02418}).
    (c) Slice-wise training-free approaches that rely on generic 2D foundation models and random feature projection (\eg RAPTOR~\citep{DBLP:conf/icml/AnJLGYS25}).
    (d) Our lightweight MVSC framework, which learns a visual surrogate to enable lightweight and effective feature extraction with generic 2D foundation models (\eg DINOv3~\citep{DBLP:journals/corr/abs-2508-10104}). 
    The reported performance are achieved on the largest AD classficiation dataset ADNI.
    }
    \label{fig:intro}
\end{figure}

Alzheimer’s disease (AD) is a progressive neurodegenerative disorder that affects millions of individuals worldwide~\citep{Chen2024-qb}. 
The pathological progression of AD is closely associated with structural brain changes, particularly the atrophy of grey matter in regions such as the hippocampus and cerebral cortex, as well as the enlargement of ventricular spaces~\citep{Frisoni2010-vp}. 
Structural magnetic resonance imaging (sMRI) is the commonly used screening technique by doctors for visualizing and checking these brain regions~\citep{DBLP:journals/cmpb/BeheshtiDFYMI16, Popuri2020-fa}.

Recent studies have proposed various techniques to learn representations from sMRI for AD classification.
Due to the 3D nature of sMRI data,
many of them utilize 3D neural network architectures, such as inflated CNN backbones ~\citep{DBLP:conf/wacv/ChenFBH24, DBLP:conf/cvpr/JangH22} (Figure~\ref{fig:intro} (a)). 
Although these 3D models demonstrate strong performance, they are computationally expensive, which limits their scalability and practical deployment.
To reduce computational cost, some methods adopt a slice-wise learning paradigm (Figure~\ref{fig:intro} (b)), where features are first learned independently from each slice of sMRI and then aggregated using trainable fusion modules ~\citep{DBLP:journals/corr/abs-2407-02418, DBLP:conf/miccai/DiaoFYWTHYX25}. 
However, independently learned slice representation ignores long-range cross-slice dependencies, which limits the fusion module’s ability to consider global structure.
Another line of research utilizes generic 2D foundation models pretrained on large-scale datasets for sMRI feature extraction. (Figure~\ref{fig:intro}(c))~\citep{DBLP:journals/corr/abs-2510-23415, DBLP:journals/corr/abs-2509-21249, DBLP:conf/icml/AnJLGYS25}.
Due to extensive pretraining and strong generalization, these models learn universal representations that capture diverse structural patterns. This results in more discriminative features that benefit classification. 
However, due to the lack of task-specific optimization, 
directly applying these models cannot fully realize their potentials.

In this paper, we propose Multimodal Visual Surrogate Compression (MVSC) (Figure~\ref{fig:intro}(d)), an adaptive and lightweight framework that bridges 3D sMRI data with 2D vision foundation models for Alzheimer’s disease classification. MVSC compresses a 3D sMRI volume into a 2D visual surrogate, which is a compact 2D feature. Specifically, we compress three 3D sMRI transformations: intensity-normalized slices, tissue segmentations, and brain masks, each obtained from the same sMRI volume, resulting in three corresponding 2D features. Together, these features form a three-channel visual surrogate that naturally aligns with the input format of generic 2D vision foundation models.

Our MVSC framework comprises two main components: a Volume Context Encoder (VoCE) and an Adaptive Slice Fusion (ASF) module. The input sMRI volume is decomposed into a set of 2D patches and embedded using a 2D CNN-based patch encoder. 
By operating at the patch-level,
MVSC allows local discriminative patterns to be directly captured and integrated to the final visual surrogate. 
However, relying solely on cross-patch relationships within individual slices and same location patch interactions across slices fails to capture global volumetric dependencies. 
%
%
%
Therefore, we equip the VoCE module with the ability to learn global context from all slices of a volume. It leverages observation-based medical text generated by LLaVA-Med~\citep{DBLP:conf/nips/LiWZULYNPG23} to guide global feature learning. The text provides semantic cues that reduce sensitivity to screening protocol variations and anatomical diversity. VoCE aggregates visual features from all slices into a global feature.
%
%

The ASF module of MVSC performs patch-level multimodal integration and cross-slice feature fusion. To enrich patch-level features with slice-specific semantics, slice-level text generated by LLaVA-Med is integrated into each patch. Using the global feature produced by VoCE, ASF applies cross-attention to fuse patch-aligned\footnote{Patch-aligned refers to patches at the same spatial location across~slices.} features across slices, producing a unified surrogate feature for the sMRI volume. The resulting patch-level surrogate feature is reshaped and decoded into a three-channel visual surrogate via a transposed 2D CNN.
The visual surrogates are then fed to a generic 2D vision foundation model to extract discriminative features, which are used for final classification by a lightweight MLP. 
By minimizing the classification loss, MVSC learns to construct 2D visual surrogate representations that are well-suited for generic 2D foundation models.

\begin{figure*}[h]
    \centering
        \includegraphics[
        trim=26cm 38cm 28cm 3cm,
        clip,
        width=1\linewidth
    ]{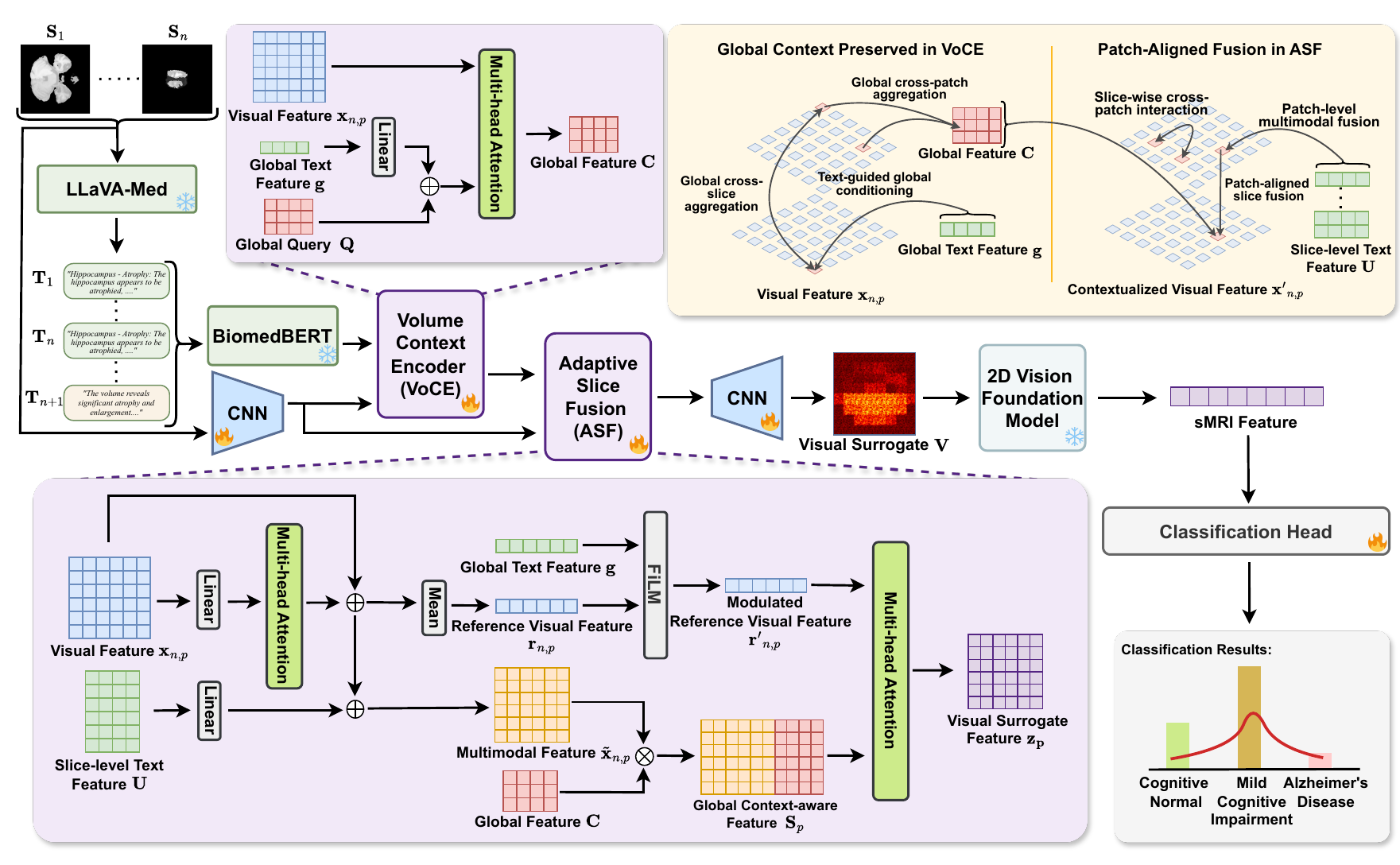}
    \caption{Main architecture of our multimodal visual surrogate compression (MVSC) framework. MVSC consists of two key components: (i) Volume Context Encoder (VoCE) (Section~\ref{subsec:voce}), which aggregates patch-level visual features across all slices to learn a global volume representation guided by volume-level text; and (ii) Adaptive Slice Fusion (ASF) (Section~\ref{subsec:asf}), which performs patch-aligned cross-slice fusion by integrating patch-level visual features with slice-level semantic information and the global representation to construct a 2D visual surrogate.}
    \label{fig:pipeline}
    \vspace{-0.2cm}
\end{figure*}



The main contributions of this paper are summarized as follows: 
\begin{itemize}
    \vspace{-0.35cm}
    \item We propose an adaptive, lightweight framework, Multimodal Visual Surrogate Compression (MVSC), which enables generic 2D foundation models to effectively extract discriminative features from 3D sMRI
    for AD classification.
    \vspace{-0.15cm}
    \item We design a Volume Context Encoder that captures essential global volumetric dependencies within 3D sMRI volumes guided by global text representation, and an Adaptive Slice Fusion module that adaptively compresses multimodal 3D representations into compact 2D features. 
    \vspace{-0.15cm}
    \item Extensive experiments demonstrate that MVSC performs favourably compared to several state-of-the-art methods in model performance.
\end{itemize}

\section{Related Work}

\paragraph{Learning on sMRI.}

Thanks to the strong capability of sMRI to capture disease-related patterns, many works have been proposed to learn on sMRI for various neuroscience tasks. For example, BrainHarmonix~\citep{dong2025brain} presents a large-scale brain foundation model to compress neuroimaging data (\eg sMRI and finctional sMRI) into 1D roken representations. 
The overall design emphasizes producing a compact latent space suitable for diverse downstream neuroscience tasks (\eg disorder classification). 
Raptor~\citep{DBLP:conf/icml/AnJLGYS25} tackles the practical bottleneck of learning from 3D volumes (\eg 3D sMRI) by proposing a train-free pipeline that converts a 3D volume into compact embeddings using a frozen 2D foundation model. TomoGraphView~\citep{kiechle2025tomographview3dmedicalimage} performs omnidirectional slicing on sMRI by sampling cross-sections from many viewpoints defined by points on a sphere surrounding the volume, achieving multi-view feature learning on sMRI. DinoAtten3D~\citep{DBLP:journals/corr/abs-2509-12512} proposes a lightweight method to exploit 2D self-supervised priors (DINOv2) for 3D sMRI by treating a volume as a set of 2D slices. Despite their impressive performance, existing methods struggle to effectively learn on 3D sMRI. To address this challenge, we propose an adaptive, lightweight framework, termed MVSC.

\paragraph{Alzheimer's Disease Diagnosis.}

Many studies have proposed state-of-the-art machine learning techniques for Alzheimer’s disease (AD) diagnosis~\citep{yang2023mapping,bian2023adversarially,jones2022computational}. 
MGCA-RAFFNet~\citep{10294288} focuses on distinguishing between different AD stages, proposing a cross-attention region-aware feature fusion network with various brain region templates. Cross-GNN~\citep{yang2023mapping} uses cross-modal mutual learning to complete binary classification tasks in AD detection. 
AGT~\citep{cho2024neurodegenerative} introduces a node-variant convolution mechanism that adaptively captures both localized homophily and heterophily to learn the progressive degenerative changes in the brain, enabling the distinction of AD stages. CSWCL-GCNs~\citep{hu2024cost} is a contrastive learning framework based on functional connectivity (FC) and high-order functional connectivity (HOFC) to address class imbalance in diagnosing different AD stages. 

\paragraph{Medical Foundation Model.}

Recent years have witnessed rapid progress in medical foundation models, which leverage large-scale medical data and pretraining strategies to support a wide range of downstream medical tasks~\cite{DBLP:conf/emnlp/0008WA022,boecking2022making,moor2023foundation,tham2025building,bannur2023learning,wu2025towards}. For example, MedUnifier~\cite{zhang2025medunifier} proposes a vision-language pretraining framework for medical data, integrating text-grounded image generation capabilities with multimodal learning strategies. VISTA3D~\cite{he2025vista3d} presents a unified foundation model to achieve 3D promptable automatic segmentation and interactive editing for 3D neuroimaging. LLaVA-Med~\cite{DBLP:conf/nips/LiWZULYNPG23} aligns medical images with textual data and leverages prompts for fine-tuning, enabling the model to comprehend complex medical imaging and perform reasoning.  


\section{Method}

This section introduces our proposed Multimodal Visual Surrogate Compression (MVSC), as illustrated in Figure~\ref{fig:pipeline}. We first preprocess sMRI using a standard pipeline (details are discussed in Appendix~\ref{appendix:preprocess}). MVSC takes preprocessed 3D sMRI transformations and corresponding text features as inputs.
Each sMRI slice is partitioned into non-overlapping patches and embedded using a 2D CNN-based patch encoder.
In parallel, observation-based medical text is generated using LLaVA-Med at both the slice level and the volume level.
The slice-level text describes anatomical and pathological patterns for individual slices, whereas the volume-level
text summarizes global characteristics of the entire sMRI volume (details of the medical text generation process are
provided in Appendix~\ref{appendix:text}).
Together with the volume-level text features, the patch embeddings are processed by the Volume Context Encoder (VoCE)
to produce a global feature of the sMRI volume.
The Adaptive Slice Fusion (ASF) module then performs patch-aligned cross-slice feature fusion, integrating patch-level visual features
with slice-level text features and the previously produced global feature to construct a compact 2D visual surrogate feature.
The fused patch features are reshaped into a compact feature, which is then decoded by a transposed 2D CNN into a three-channel visual surrogate. This visual surrogate is fed to a generic pretrained 2D vision foundation model for feature extraction, followed by AD classification. Below, we provide details for each of the main components.



%

%


\noindent\textbf{Preliminaries.}
In this paper, we define a visual surrogate as a three-channel 2D representation compressed from a 3D sMRI volume. Each channel corresponds to a distinct transformation of the sMRI data (\eg intensity-normalized slices, tissue segmentations, or brain masks).

\paragraph{Notations.} 
Index sets are written as $[N]=\{1,\dots,N\}$. Scalars are denoted by regular fonts (\eg $n, p, d$), vectors by lowercase bold letters (\eg $\mathbf{x}$); matrices by uppercase bold letters (\eg $\mathbf{X}$); and higher-order tensors by calligraphic letters (\eg $\mathcal{X}$).

\subsection{Volume Context Encoder (VoCE)}
\label{subsec:voce}




sMRI volumes are high-dimensional and structurally complex, consisting of a large number of slices that encode region-specific tissue information at different positions across the brain. Neighboring slices exhibit strong spatial overlap, while distant slices provide higher-level contextual information. Together, neighboring and distant slices contribute to the global volumetric dependencies, making it essential to model context across the entire volume.

To capture global volumetric dependencies prior to fine-grained slice fusion, we introduce the VoCE. VoCE aggregates visual information across all slices and patches into a global feature, guided by global text features.


\paragraph{Global Feature Aggregation.}
Let the sMRI visual feature patches encoded by 2D CNN be denoted as
$\{\mathbf{x}_{n,p}\}_{n \in [N],\, p \in [P]}$, where $n$ indexes slices and $p$ indexes patches. In addition, we obtain the text from LLaVA-Medv1.5 ~\citep{li2023llava} and its embeddings
$\mathbf{T} \in \mathbb{R}^{(N+1) \times d_t}$, consisting of $N$ slice-level and $1$ global text representation, from BiomedBERT~\citep{DBLP:journals/health/GuTCLULNGP22}. The text embeddings are projected to the feature space with the dimension $d$ via a linear map $\phi(\cdot)$, resulting a global text feature $\mathbf{g} \in \mathbb{R}^{d}$.

VoCE learns a set of $K$ global query tokens,
$\mathbf{Q} \in \mathbb{R}^{K \times d}$, for storing global volumetric information. Then, we inject global text guidance
into each vector,
\begin{equation}
\mathbf{L} = \mathbf{Q} + \mathbf{g},
\end{equation}
where $\mathbf{L} \in \mathbb{R}^{K \times d}$ denotes the text-guided queries.

The queries $\mathbf{L}$ attend over all visual tokens in the volume using multi-head cross-attention~\citep{DBLP:conf/nips/VaswaniSPUJGKP17}:
\begin{equation}
\mathbf{C} = \mathrm{Attn}(\mathbf{L}, \{\mathbf{x}_{n,p}\}),
\end{equation}
where $\mathbf{C} \in \mathbb{R}^{K \times d}$ is the resulting set of global features.
Each vector in $\mathbf{C}$ aggregates information from all slices, forming a compact global representation of the volume. VoCE provides a text-guided summary of the volume that can be used in downstream ASF module.


\subsection{Adaptive Slice Fusion (ASF)}
\label{subsec:asf}

Given the set of global features ($\mathbf{C}$) extracted by VoCE, ASF integrates it with patch-level visual features, slice-level text features to perform patch-aligned cross-slice fusion, resulting a visual surrogate feature. Specifically, the fusion is guided by the global text and slice-averaged reference token.






\paragraph{Patch-level Feature Augmentation.}

We apply self-attention $\mathrm{SA}(\cdot)$ over patches to model cross-patch interactions within each slice:
\begin{equation}
\mathbf{X}_n' = \mathrm{SA}(\mathbf{X}_n),
\end{equation}
where $\mathbf{X}_n = (\mathbf{x}_{n,1}, \ldots, \mathbf{x}_{n,P}) \in \mathbb{R}^{P \times d}$ denotes the patch-level visual features of the $n$-th slice. The output $\mathbf{X}_n'$ represents contextualized patch features after self-attention. We denote the $p$-th token of $\mathbf{X}_n'$ by $\mathbf{x}'_{n,p}$.

To further enhance patch-level features with slice-level semantic information, we incorporate slice-level text features into the patch-level visual features. Each slice-level text feature $\mathbf{u}_n$ is projected into the visual feature space and added to the corresponding patch features (all patches within the same slice share the same slice-level text):
\begin{equation}
\tilde{\mathbf{x}}_{n,p} = \mathbf{x}'_{n,p} + \psi(\mathbf{u}_n),
\end{equation}
where $\psi(\cdot)$ denotes a learnable linear projection.

In addition, we incorporate global features $\mathbf{C} \in \mathbb{R}^{K \times d}$ produced by VoCE as additional global information. For each patch index $p \in [P]$, we form a set of global-aware features by concatenating the slice-level visual features at that patch index with the global features:
\begin{equation}
\mathcal{S}_p
=
(\tilde{\mathbf{x}}_{1,p}, \ldots, \tilde{\mathbf{x}}_{N,p}, \mathbf{c}_1, \ldots, \mathbf{c}_K)
\in \mathbb{R}^{(N+K)\times d}.
\end{equation}

\paragraph{Patch-aligned Cross-slice Fusion.}
We compute a slice-averaged reference token at each patch index to provide a shared patch-level reference for consistent cross-attention across slices.
Specifically, for each patch index $p \in [P]$, the reference token $\mathbf{r}_p \in \mathbb{R}^{d}$ is obtained by averaging the corresponding visual tokens across slices:
\begin{equation}
\mathbf{r}_p = \frac{1}{N} \sum_{n \in [N]} \mathbf{x}'_{n,p}.
\end{equation}


To inject global text information into the slice-averaged reference tokens via feature-wise modulation, we apply FiLM~\citep{DBLP:conf/aaai/PerezSVDC18} using the global text feature $\mathbf{g}$:
\begin{equation}
\mathbf{r}_p' = \gamma(\mathbf{g}) \odot \mathbf{r}_p + \beta(\mathbf{g}),
\end{equation}
where $\gamma(\cdot)$ and $\beta(\cdot)$ are learnable linear projections that produce feature-wise scale and bias, respectively. The resulting vector $\mathbf{r}_p' \in \mathbb{R}^{d}$ is referred to as the modulated reference token and serves as the query for slice-level fusion at patch index $p$.



Then, for each patch index $p \in [P]$, we apply multi-head cross-attention using the modulated reference token $\mathbf{r}_p'$ as the query and the corresponding feature set $\mathcal{S}_p$ as keys and values:
\begin{equation}
\mathbf{z}_p = \mathrm{Attn}(\mathbf{r}_p', \mathcal{S}_p).
\end{equation}
%
The resulting fused tokens $\{\mathbf{z}_p\}_{p \in [P]}$ encode patch-level representations that integrate patch visual information, slice-level visual and text context, and global volumetric features at each patch index.

The fused tokens $\{\mathbf{z}_{p}\}_{p \in [P]}$ are reshaped into a 2D representation $\mathbf{Z}$ and decoded via a transposed convolution to produce a three-channel visual surrogate $\mathbf{V}$.

\subsection{Classification}
The visual surrogate $\mathbf{V}$ generated by ASF is then processed by a frozen 2D vision foundation model to extract high-level features, which are subsequently used for AD classification by a lightweight three-layer fully connected network.
%



\section{Experiments}

We evaluate MVSC on the task of Alzheimer’s disease classification, considering both binary classification and multi-class classification. The comparison is made against a diverse set of existing methods. In addition, we assess MVSC across multiple frozen 2D vision foundation models and backbone sizes to evaluate its scalability.

\subsection{Set Up}


\paragraph{Datasets.}

Inspired by~\citet{DBLP:conf/miccai/PuglisiAR24}, we evaluate our method on three widely used sMRI Alzheimer’s disease datasets. \textbf{(i)} AIBL~\citep{Ellis2009-fd} (Australian Imaging, Biomarker \& Lifestyle) is a longitudinal study designed to investigate cognitive aging and Alzheimer’s disease. The dataset includes 790 sMRI scans from 242 individuals, comprising 596 cognitively normal (CN), 105 mild cognitive impairment (MCI), and 89 Alzheimer’s disease (AD) cases. \textbf{(ii)} OASIS-3~\citep{LaMontagne2019.12.13.19014902} (Open Access Series of Imaging Studies) provides imaging and clinical data spanning normal aging to dementia. We use 1,662 sMRI scans from 567 individuals, including 1,231 CN, 58 MCI, and 373 AD cases. \textbf{(iii)} ADNI~\citep{Petersen2010-gf} (Alzheimer’s Disease Neuroimaging Initiative) is the largest public dataset for Alzheimer’s disease research. We combine sMRI data from ADNI-1, ADNI-2, ADNI-3, and ADNI-GO, resulting in 9,266 sMRI scans from 1,990 individuals, including 3,421 CN, 4,825 MCI, and 1,020 AD cases.

All sMRI volumes are preprocessed using a standard pipeline that includes N4 bias-field correction~\citep{DBLP:journals/tmi/TustisonACZEYG10}, skull stripping~\citep{DBLP:journals/neuroimage/HoopesMDFH22}, affine registration to MNI space, intensity normalization~\citep{Shinohara2014-vx}, and resampling to a voxel spacing of $1.0\,\mathrm{mm} \times 1.0\,\mathrm{mm} \times 1.0\,\mathrm{mm}$. Details of the preprocessing procedure are provided in Appendix~\ref{appendix:preprocess}.



\paragraph{Implementation.}
Due to the large number of slices in each sMRI volume, we select a fixed number of representative slices (\eg 10, 20, 50, or 100) based on intensity statistics (details discussed in Appendix~\ref{appendix:selection}), rather than using the full volume. Medical text for the selected slices are generated using LLaVA-Medv1.5~\citep{DBLP:conf/nips/LiWZULYNPG23} and encoded with BiomedBERT~\citep{DBLP:journals/health/GuTCLULNGP22}. 

We employ a lightweight MLP classification head optimized with cross-entropy loss and AdamW. The model is trained with a learning rate of 0.0001, 10 warmup epochs followed by cosine annealing over 150 epochs, and a weight decay of 0.0001. MVSC has a compact model size, highlighting its lightweight design. Specifically, MVSC uses 9.81M parameters on ADNI, while a smaller configuration with 0.7M parameters is adopted for the smaller AIBL and OASIS datasets. Additional implementation details and training settings are provided in Appendix~\ref{appendix:implementation}. The code and full implementation will be released.




\begin{table*}[ht]
\centering
\tiny 
\renewcommand\arraystretch{0.6}
 \setlength{\tabcolsep}{6pt} 
\caption{Experimental results on binary AD classification (CN vs. AD) across AIBL, ADNI, and OASIS datasets. The best results are marked in \textbf{bold}, and the suboptimal results are marked \underline{underlined}.}

\begin{adjustbox}{width=1\textwidth}
\begin{tabular}{lcccccc}
    \toprule
    \multirow{2}{*}{Method} 
        & \multicolumn{2}{c}{AIBL} 
        & \multicolumn{2}{c}{ADNI} 
        & \multicolumn{2}{c}{OASIS}\\
    \cmidrule(lr){2-3}
    \cmidrule(lr){4-5}
    \cmidrule(lr){6-7}
     & AUC  & Acc 
     & AUC  & Acc
     & AUC  & Acc\\
    \midrule
    3D ResNet        
        & 93.95\%  & 89.93\%
        & 92.02\% & 89.17\%
        & 80.22\% & 76.71\% \\
    M3T  
        & 94.96\% & \underline{95.07\%}
        & 88.83\% & 86.03\%
        & 70.10\% & 77.09\% \\
    MERLIN   
        & 94.49\%  &  91.37\%
        & 93.42\% & 90.77\%
        & \underline{85.88\%}  & \underline{85.09\%} \\
    \midrule
    ViT       
        & 95.75\%  & 93.53\%
        & 90.92\% & 86.49\%
        & 85.47\% & 83.58\% \\
    PPAL  
        & 61.89\% & 90.47\%
        & 79.95\% & 86.08\%
        & 72.35\% & \textbf{85.71\%}\\
    AXIAL  
        & 51.92\%  & 81.34\% 
        & 84.73\%  & 85.92\%
        & 57.61\%  & 65.15\% \\
    \midrule
    RAPTOR  
        & 95.06\%  & 94.96\%
        & 92.95\% & 89.86\%
        & 77.72\% & 80.12\% \\
    \midrule
    Mean (DINOv3)       
        & 83.28\%  & 82.73\%
        & 65.77\% &  77.13\%
        & 69.54\% &  73.60\% \\
    Max (DINOv3)       
        & 71.79\%  & 83.45\%
        & 55.68\% & 75.45\%
        & 56.91\% &  73.91\% \\
    Mean (I-JEPA)       
        & 91.94\%  & 90.53\%
        & 90.29\% &  85.02\%
        & 80.87\% &  85.01\% \\
    Max (I-JEPA)       
        & 88.90\%  & 80.57\%
        & 73.63\% & 77.27\%
        & 79.14\% &  83.22\% \\
    \midrule
    MVSC (Ours)(DINOv3)        
        & \underline{97.19\%}  & \textbf{95.68\%}
        & \underline{96.08\%} & \underline{91.89\%}
        & 84.79\% & 83.54\% \\
    MVSC (Ours)(I-JEPA)        
        & \textbf{98.50\%} & 93.52\%
        & \textbf{97.16\%} & \textbf{93.35\%}
        & \textbf{86.41\%} & 83.85\% \\
    \bottomrule
\end{tabular}
\end{adjustbox}
\label{tab:main_results}
\end{table*}

\paragraph{Metrics.}
We adopt the area under the receiver operating characteristic curve (AUC) as the primary evaluation metric, as Alzheimer’s disease datasets are inherently imbalanced, with more cognitively normal (CN) and mild cognitive impairment (MCI) cases than Alzheimer’s disease (AD) cases. For multi-class classification, we report the macro-averaged AUC (mAUC). 
In addition, we report classification accuracy as a complementary metric.

\begin{table*}[tbp]
\centering
\renewcommand\arraystretch{0.6}
 \setlength{\tabcolsep}{11pt} 
\caption{Experimental results on multi-class AD classification (CN vs. MCI vs. AD) across AIBL, ADNI, and OASIS datasets.}

\begin{adjustbox}{width=1\textwidth}
\begin{tabular}{lcccccc}
    \toprule
    \multirow{2}{*}{Method} 
        & \multicolumn{2}{c}{AIBL} 
        & \multicolumn{2}{c}{ADNI} 
        & \multicolumn{2}{c}{OASIS}\\
    \cmidrule(lr){2-3}
    \cmidrule(lr){4-5}
    \cmidrule(lr){6-7}
     & mAUC  & Acc 
     & mAUC  & Acc
     & mAUC  & Acc\\
    \midrule
    3D ResNet  
        & \underline{85.16}\% & 80.51\%
        & 65.90\% & 52.75\%
        & 66.49\% & 73.73\% \\
    MERLIN 
        & 80.91\% & \underline{81.99}\%
        & 89.70\% & 78.81\%
        & \underline{82.97\%} & \textbf{82.69\%} \\
    \midrule
    ViT  
        & 67.69\% & 74.33\%
        & 60.32\% & 52.08\%
        & 72.20\% & 73.73\% \\
    \midrule
    RAPTOR
        & 82.84\% & 80.12\%
        & 80.05\% & 66.52\%
        & 80.81\% & 76.41\% \\
    \midrule
    MVSC (Ours)(DINOv3)        
        & \textbf{85.42\%}  & \textbf{83.23\%}
        &  \underline{93.05\%} & \underline{82.15\%}
        &  82.31\% & \underline{79.40\%} \\
    MVSC (Ours)(I-JEPA)        
        & 84.73\%  & 81.36\%
        & \textbf{93.65\%}&  \textbf{83.82}\%
        & \textbf{83.01\%} & 75.82\% \\
    \bottomrule
\end{tabular}
\end{adjustbox}
\label{tab:multi_results}
\end{table*}

\subsection{Results}

We evaluate MVSC paired with frozen DINOv3~\citep{DBLP:journals/corr/abs-2508-10104} and I-JEPA~\citep{DBLP:conf/cvpr/AssranDMBVRLB23} backbones. By default, DINOv3 uses a ViT-B/16 backbone, while I-JEPA uses a ViT-H/14 backbone. Unless otherwise specified, we adopt a default configuration with Top-$k = 50$. 
We compare MVSC against a diverse set of methods: (a) inflated CNN including 3D ResNet (ResNet50)~\cite{fan2021pytorchvideo}, MERLIN (image encoder only)~\cite{DBLP:journals/corr/abs-2406-06512}, and M3T~\cite{DBLP:conf/cvpr/JangH22}; (b) slice-wise learning methods (PPAL, AXIAL, and ViT (ViT-S/16))~\cite{DBLP:conf/miccai/DiaoFYWTHYX25,DBLP:journals/corr/abs-2407-02418,DBLP:conf/iclr/DosovitskiyB0WZ21}; (c) generic 2D foundation models (RAPTOR)~\cite{DBLP:conf/icml/AnJLGYS25}; (d) deterministic volumetric compression strategies (mean and max pooling). 
All methods are evaluated under the same experimental setup on the three datasets for fair comparison.





\paragraph{Binary Classification.}
Table~\ref{tab:main_results} reports results for binary classification (CN vs. AD) on three datasets. Across all datasets, both variants of MVSC (DINOv3 and I-JEPA) achieve competitive performance. On the ADNI dataset, MVSC (I-JEPA) attains the highest AUC of 97.16\%, while MVSC (DINOv3) achieves the second-highest AUC of 96.08\%. In particular, MVSC (I-JEPA) outperforms the strongest comparison method, MERLIN, by 3.74\% in AUC. On the AIBL and OASIS datasets, MVSC (I-JEPA) also achieves the highest AUCs of 98.50\% and 86.41\%, respectively, with MVSC (DINOv3) exhibiting comparable performance. These results demonstrate that MVSC enables generic 2D foundation models to effectively extract discriminative features from 3D sMRI. As a result, MVSC achieves superior performance compared to inflated CNN-based methods, slice-wise learning approaches, and generic 2D foundation models.

\begin{table*}[tbp]
\centering
\renewcommand\arraystretch{0.6}
 \setlength{\tabcolsep}{7pt} 
\caption{Class-wise AUC results for multiclass AD classification across AIBL, ADNI, and OASIS datasets. 
}
\begin{adjustbox}{width=1\textwidth}
\begin{tabular}{lccccccccc}
    \toprule
    \multirow{2}{*}{Method} 
        & \multicolumn{3}{c}{AIBL} 
        & \multicolumn{3}{c}{ADNI} 
        & \multicolumn{3}{c}{OASIS}\\
    \cmidrule(lr){2-4}
    \cmidrule(lr){5-7}
    \cmidrule(lr){8-10}
     & CN  & {MCI} & AD 
     & CN  & {MCI} & AD 
     & CN  & {MCI} & AD \\
    \midrule
    3D ResNet        
        & 88.65\%  & 79.83\% & 87.02\%
        & 63.96\%  & 58.13\% & 75.62\%
        & 66.24\%  & 65.43\% & 67.79\% \\
    MERLIN  
        & 81.83\%  & 73.32\% & 87.58\%
        & 90.64\%  & 87.65\% & 90.82\%
        & 83.22\%  & 79.27\% & 86.43\% \\
    \midrule
    ViT  
        & 69.28\%  & 57.13\% & 76.67\%
        & 58.65\%  & 49.21\% & 73.10\%
        & 74.28\%  & 71.72\% & 70.62\% \\
    \midrule
    RAPTOR        
        & 82.80\%  & 74.10\% & 91.62\%
        & 82.75\%  & 75.90\% & 81.52\%
        & 76.35\%  & \textbf{91.69\%} & 74.41\% \\
    \midrule
    MVSC (Ours)(DINOv3)        
        & 86.30\%  & \textbf{81.82\%} & 88.14\%
        & 93.85\%  & 92.18\% & 93.11\%
        & 82.81\%  & 80.43\% & 83.68\% \\
    MVSC (Ours)(I-JEPA)  
        & 84.14\%  & 77.59\% & 92.47\%
        & 94.59\%  & \textbf{93.33\%} & 93.05\%
        & 84.76\%  & 78.18\% & 86.08\% \\
    \bottomrule
\end{tabular}
\end{adjustbox}
\label{tab:multi_split_results}
\end{table*}


\paragraph{Multi-class Classification.}
We further evaluate MVSC on the more challenging multi-class classification task (CN vs. MCI vs. AD), with results reported in Table~\ref{tab:multi_results}. Overall, both variants of MVSC achieve strong performance across all three datasets. In particular, MVSC (I-JEPA) attains the highest mAUC of 93.65\% and 83.01\% on ADNI and OASIS, respectively, while MVSC (DINOv3) achieves the highest mAUC of 85.42\% on AIBL. Notably, MVSC (I-JEPA) outperforms the strongest comparison method, MERLIN, by 3.95\% on ADNI. These results demonstrate that even under more challenging multi-class settings, MVSC consistently delivers superior performance compared to inflated CNN-based methods, slice-wise learning approaches, and generic 2D foundation models across diverse datasets.

Among the three classes in AD classification, MCI is generally considered the most challenging class to distinguish, even for human experts~\cite{mci}. Therefore, we further report class-wise AUC results to analyze model performance on the MCI class. As shown in Table~\ref{tab:multi_split_results}, although MVSC achieves the second-highest AUC for MCI on OASIS, it attains the best AUC for MCI on both ADNI and AIBL. Specifically, on ADNI, MVSC (I-JEPA) achieves the highest AUC for the MCI class at 93.33\%, outperforming the strongest comparison method, MERLIN, by 5.68\%, while MVSC (DINOv3) attains the second-highest AUC of 92.18\%. On AIBL, MVSC (DINOv3) achieves the highest AUC for the MCI class at 81.82\%, with MVSC (I-JEPA) exhibiting comparable performance. These results show that MVSC effectively preserves fine-grained pathological patterns in challenging multi-class scenarios, leading to improved classification performance.

\subsection{Ablation Study}

In this section, we conduct ablation studies to evaluate the contribution of VoCE by removing it and relying solely on cross-patch relationships within slices. In contrast, an ablation study of the ASF module is not feasible, as ASF constructs the visual surrogate feature required by downstream 2D foundation models. We additionally conduct an ablation study on the medical text modality to examine its impact on downstream classification performance.

\vspace{-0.3cm}
\paragraph{Contribution of VoCE.}      
To assess the contribution of the VoCE, we conduct ablation studies on both binary and multi-class classification tasks. 
Results are presented in Table~\ref{tab:ablation}, denoted as `w/o VoCE'. We first compare the full MVSC model (VoCE+ASF) against a variant without VoCE, where the global context features are removed, and the cross-attention in ASF operates solely on local patch features within each slice.

Across all datasets, removing VoCE causes a consistent performance drop. For binary classification, the full model outperforms the w/o VoCE variant on all three datasets, demonstrating the importance of introducing global volumetric context. The effect is even more pronounced in the multi-class setting, where distinguishing between adjacent disease stages requires richer global information. In particular, on ADNI, the full model improves mAUC by 9.44\% and accuracy by 12.99\% over the w/o VoCE variant.
This gap highlights the role of VoCE in capture global volumetric dependencies that are especially critical for augmenting discriminative features in diverse datasets.

\begin{table}[tbp]
    \centering
    \tiny
    \caption{Ablation study across AIBL, ADNI, and OASIS, evaluating the impact of global context encoding (VoCE) and textual conditioning on binary and multiclass AD classification.}
    \begin{tabularx}{\columnwidth}{l *{6}{c}}
        \toprule
        Model Variant
        & \multicolumn{2}{c}{AIBL}
        & \multicolumn{2}{c}{ADNI}
        & \multicolumn{2}{c}{OASIS} \\
        \cmidrule(lr){2-3}
        \cmidrule(lr){4-5}
        \cmidrule(lr){6-7}
         & (m)AUC & Acc
         & (m)AUC & Acc
         & (m)AUC & Acc \\
        \midrule
        \multicolumn{7}{l}{\textbf{Binary}} \\
        w/o VoCE   & 92.76\% & 90.64\% & 92.07\% & 87.27\% & 82.12\% & 81.67\% \\
        w/o Text   & 92.28\% & 93.52\% & 89.19\% & 84.57\% & 84.41\% & 81.36\% \\
        VoCE+ASF  & 97.19\% & 95.68\% & 96.08\% & 91.89\% & 84.79\% & 83.54\% \\
        \midrule
        \multicolumn{7}{l}{\textbf{Multi-class}} \\
        w/o VoCE   & 81.36\% & 80.65\% & 83.61\% & 69.16\% & 78.70\% & 75.52\% \\
        w/o Text   & 81.52\% & 79.50\% & 86.82\% & 72.50\% & 80.35\% & 75.52\% \\
        VoCE+ASF  & 85.42\% & 83.23\% & 93.05\% & 82.15\% & 82.31\% & 79.40\% \\
        \bottomrule
    \end{tabularx}
    \label{tab:ablation}
    \vspace{-0.5cm}
\end{table}

\paragraph{Contribution of Medical Text.}
We assess the contribution of the medical text modality by removing it from the model. We use the same experimental configuration as the w/o VoCE variant. Results are also presented in Table~\ref{tab:ablation}, denoted as `w/o Text'. 
For both binary and multi-class tasks, the w/o Text variant shows a performance decrease compared to the full MVSC model across all three datasets.
Particularly, on ADNI, removing text leads to a noticeable performance degradation, with binary AUC dropping by 6.89\% and multi-class mAUC by 6.23\%, along with significant drops in accuracy. 
This indicates that semantic information helps the model capture discriminative features across various screening protocols and anatomical variations. 

\subsection{Scaling Study}

To assess the scalability of MVSC, we conduct scaling experiments along two axes:
\textbf{(i)} backbone architecture, covering different design families and parameter scales, and
\textbf{(ii)} input coverage, controlled by the number of representative sMRI slices (Top-$k$) used to construct the visual surrogate.
Our goal is to evaluate whether MVSC maintains competitive performance across different backbone capacities and slice coverage levels.

\begin{table}[ht]
    \centering
    \tiny
    \caption{Scaling study across AIBL, ADNI, and OASIS, evaluating the impact of backbone
    and parameter size on binary AD classification using MVSC with different backbones of DINOv3.}
    \setlength{\tabcolsep}{5pt}
    \begin{tabularx}{\columnwidth}{l *{6}{c}}
        \toprule
        \multirow{2}{*}{Backbone} 
            & \multicolumn{2}{c}{AIBL} 
            & \multicolumn{2}{c}{ADNI} 
            & \multicolumn{2}{c}{OASIS} \\
        \cmidrule(lr){2-3}
        \cmidrule(lr){4-5}
        \cmidrule(lr){6-7}
         & AUC & Acc 
         & AUC & Acc
         & AUC & Acc \\
        \midrule
        ViT-B/16 {[86M]}
            & 97.19\% & 95.68\%
            & 96.08\% & 91.89\%
            & 84.79\% & 83.54\% \\
        ViT-L/16 {[300M]}
            & 98.46\% & 96.40\%
            & 92.46\% & 88.28\%
            & 84.31\% & 83.22\% \\
        ViT-H/16 {[840M]}
            & 93.46\% & 89.20\%
            & 91.18\% & 88.61\%
            & 82.27\% & 82.91\% \\
        \midrule
        ConvNeXt-B {[89M]}
            & 96.27\% & 92.80\%
            & 92.09\% & 88.37\%
            & 87.02\% & 83.85\% \\
        ConvNeXt-L {[198M]}
            & 97.54\% & 92.08\%
            & 90.11\% & 87.04\%
            & 80.56\% & 79.50\% \\
        \bottomrule
    \end{tabularx}
    \label{tab:scaling_backbone}
\end{table}

\vspace{-0.3cm}
\paragraph{Scaling Across Backbone Architectures.}

As reported in Table~\ref{tab:scaling_backbone}, MVSC demonstrates consistent performance across two widely adopted backbone families and multiple model sizes.
Specifically, within the DINOv3 ViT family, ViT-B/16 and ViT-L/16 exhibit a favorable trade-off between model capacity and performance, achieving the highest AUC and accuracy on AIBL and maintaining competitive results on ADNI and OASIS. ViT-H/16 shows a slight performance drop. This may be attributed to its large capacity (840M parameters), which can increase the risk of overfitting when training data are limited.
For the ConvNeXt based variants~\cite{DBLP:conf/cvpr/WooDHC0KX23}, ConvNeXt-B and ConvNeXt-L also achieve competitive performance across all three datasets. In particular, ConvNeXt-B gains the highest AUC of 87.02\% on OASIS. Notably, ConvNeXt-L exhibits slightly reduced performance on ADNI and OASIS compared to other models. This may be because larger ConvNeXt models are sensitive to fine-grained spatial details. 

Overall, these results demonstrate that MVSC performs consistently with respect to backbone choice. Moreover, it can effectively adapt the learned visual surrogate across architectures and scales, even when backbone capacity varies by nearly an order of magnitude.

\paragraph{Scaling with Slice Coverage.}
Table~\ref{tab:scaling_topk} reports binary (CN vs. AD) classification performance as the number of selected sMRI slices (Top-$k$) increases from 10 to 100. 
In general, MVSC performs best when $k$ is set to 20 or 50, with OASIS being an exception where performance peaks at $k=10$. Setting $k$ to 100 does not consistently result in further improvements. For example, on AIBL, increasing $k$ from 20 to 50 leads to AUC gains of 3.46\% for I-JEPA, whereas further increasing $k$ to 100 results in a moderate decrease in AUC. On AIBL and ADNI, MVSC achieves comparable performance across increasing $k$ values with both DINOv3 and I-JEPA backbones.

\begin{table}[ht]
    \centering
    \tiny
    \caption{Scaling study across AIBL, ADNI, and OASIS, evaluating the effect of slice coverage (Top-$k$) on binary AD classification using MVSC with different frozen 2D foundation models.}
    \setlength{\tabcolsep}{5pt}
    \begin{tabular}{llcccccc}
        \toprule
        \multirow{2}{*}{Backbone} & \multirow{2}{*}{Top-$k$}
            & \multicolumn{2}{c}{AIBL}
            & \multicolumn{2}{c}{ADNI}
            & \multicolumn{2}{c}{OASIS} \\
        \cmidrule(lr){3-4}
        \cmidrule(lr){5-6}
        \cmidrule(lr){7-8}
         &  & AUC & Acc
              & AUC & Acc
              & AUC & Acc \\
        \midrule
        \multirow{3}{*}{DINOv3}
        & 10
            & 96.11\% & 95.67\%
            & 93.17\% & 91.10\%
            & 86.24\% & 85.09\% \\
        & 20
            & 97.59\% & 95.67\%
            & 94.34\% & 89.63\%
            & 82.08\% & 81.06\% \\
        & 50
            & 97.19\% & 95.68\%
            & 96.08\% & 91.89\%
            & 84.79\% & 83.54\% \\
        & 100
            & 96.14\% & 93.52\%
            & 95.13\% & 90.87\%
            & 79.88\% & 79.50\% \\
        \midrule
        \multirow{3}{*}{I-JEPA}
        & 10
            & 94.91\% & 92.80\%
            & 96.94\% & 92.79\%
            & 88.38\% & 85.11\% \\
        & 20
            & 95.04\% & 94.96\%
            & 96.25\% & 92.34\%
            & 82.64\% & 83.22\% \\
        & 50
            & 98.50\% & 93.52\%
            & 97.16\% & 93.35\%
            & 86.41\% & 83.85\% \\
        & 100
            & 95.26\% & 93.52\%
            & 96.80\% & 93.35\%
            & 86.29\% & 85.70\% \\
        \bottomrule
    \end{tabular}
    \label{tab:scaling_topk}
\end{table}

\begin{table}[!ht]
    \centering
    \tiny
    \caption{Scaling study across AIBL, ADNI, and OASIS, evaluating the effect of slice coverage
    (Top-$k$) on multiclass AD classification using MVSC with different frozen 2D foundation models.}
    \setlength{\tabcolsep}{5pt}
    \begin{tabular}{llcccccc}
        \toprule
        \multirow{2}{*}{Backbone} & \multirow{2}{*}{Top-$k$}
            & \multicolumn{2}{c}{AIBL}
            & \multicolumn{2}{c}{ADNI}
            & \multicolumn{2}{c}{OASIS} \\
        \cmidrule(lr){3-4}
        \cmidrule(lr){5-6}
        \cmidrule(lr){7-8}
         &  & mAUC & Acc
              & mAUC & Acc
              & mAUC & Acc \\
        \midrule
        \multirow{3}{*}{DINOv3}
            & 10
            & 85.42\% & 80.75\%
            & 87.58\% & 76.39\%
            & 80.14\% & 77.91\% \\
        & 20
            & 82.62\% & 81.37\%
            & 89.30\% & 78.00\%
            & 82.14\% & 78.81\% \\
        & 50
            & 85.42\% & 83.23\%
            & 93.05\% & 82.15\%
            & 82.31\% & 79.40\% \\
        & 100
            & 81.06\% & 80.51\%
            & 91.28\% & 80.59\%
            & 78.65\% & 76.11\% \\
        \midrule
        \multirow{3}{*}{I-JEPA}
            & 10
            & 80.22\% & 80.74\%
            & 87.96\% & 75.41\%
            & 88.12\% & 81.19\% \\
        & 20
            & 88.38\% & 83.85\%
            & 93.96\% & 85.01\%
            & 85.42\% & 83.28\% \\
        & 50
            & 84.73\% & 81.36\%
            & 93.65\% & 83.82\%
            & 83.01\% & 75.82\% \\
        & 100
            & 88.09\% & 83.85\%
            & 92.54\% & 82.10\%
            & 82.33\% & 82.38\% \\
        \bottomrule
    \end{tabular}
    \label{tab:scaling_topk_multi}
    \vspace{-0.3cm}
\end{table}




\vspace{-0.15cm}
We further analyze the effect of slice coverage in the multi-class setting (CN vs. MCI vs. AD), with results shown in Table~\ref{tab:scaling_topk_multi}. Overall, the model performance is encouraging across different Top-$k$ values for both DINOv3 and I-JEPA backbones. Similar to the binary classification setting, MVSC achieves its best performance when $k$ is set to 20 or 50. For example, on ADNI, MVSC with the I-JEPA backbone achieves its highest mAUC of 93.96\% at $k = 20$, whereas MVSC with the DINOv3 backbone attains its best mAUC of 93.05\% at $k = 50$. An exception is the I-JEPA variant on OASIS, which performs best when $k = 10$. This may be because discriminative pathological information in OASIS is concentrated in a small set of central slices, and adding more slices introduces redundant or noisy information.



Overall, MVSC maintains competitive performance across all Top-$k$ settings. This suggests that a moderate slice coverage is sufficient to capture discriminative information.


\section{Conclusion}
In this work, we propose MVSC, an adaptive, lightweight multimodal surrogate compression framework designed to bridge sMRI and generic frozen 2D vision foundation models. By jointly modelling global volumetric context and adaptive patched-aligned slice fusion under text guidance, MVSC captures 
discriminative information
and yielding a compact 2D visual surrogate. This design reduces reliance on costly 3D architectures and enables lightweight, task-adaptive learning from high-dimensional sMRI data. 
Extensive results on multiple Alzheimer’s disease benchmarks demonstrate consistent gains in both binary and multi-class classification, underscoring the effectiveness of learned surrogate representations.






\bibliography{ref}
\bibliographystyle{icml2026}

\newpage
\appendix
\onecolumn

\section{sMRI Preprocessing}
\label{appendix:preprocess}

All sMRI volumes are preprocessed using a standard pipeline that includes N4 bias-field correction~\citep{DBLP:journals/tmi/TustisonACZEYG10}, skull stripping~\citep{DBLP:journals/neuroimage/HoopesMDFH22}, affine registration to MNI space, intensity normalization~\citep{Shinohara2014-vx}, and resampling to an isotropic voxel spacing of $1.0 \times 1.0 \times 1.0$\,mm. After preprocessing, each volume has a spatial dimension of $182 \times 218 \times 182$ voxels.

Brain tissue segmentations are then obtained using SynthSeg~2.0~\citep{DBLP:journals/mia/BillotGPTLFDI23}. Each volume is further resized to a spatial resolution of $256 \times 256$ in the axial plane. For each slice, three processed sMRI representations the intensity normalized image, the brain mask, and the tissue segmentation are stacked along the channel dimension to form a three-channel input.

\section{Medical Text Generation}
\label{appendix:text}
We generate contextual semantic descriptions from selected sMRI slices using a medical vision-language model (VLM) and then convert the generated texts into fixed-length embeddings using a biomedical text encoder.
Specifically, we use LLaVA-Med as the captioning model to produce (i) slice-level reports and (ii) a volume-level summary, and we use BiomedBERT as the text embedding model to obtain normalized text features.

\paragraph{Prompts.}
Let $\mathcal{V}\in[0,1]^{Z\times H\times W}$ be a normalized sMRI volume and let $\mathcal{I}_k=\{z_1,\dots,z_k\}\subseteq[Z]$ be the indices of the selected $k$ slices (Appendix~\ref{appendix:selection}).
For each selected slice $z\in\mathcal{I}_k$, we form an RGB-like image $\mathbf{I}_z\in\{0,\dots,255\}^{H\times W\times 3}$ by repeating the grayscale intensities across channels, and query the VLM with a fixed instruction prompt:
\[
\mathbf{c}_z \;=\; f_{\text{VLM}}\big(\mathbf{I}_z,\; p_{\text{slice}}\big), 
\qquad \forall z\in\mathcal{I}_k,
\]
where $f_{\text{VLM}}(\cdot)$ denotes the deterministic generation procedure ( \ie $\texttt{do\_sample}= \texttt{False}$), $\mathbf{c}_z$ is the generated slice caption text, and the slice prompt is
\[
\begin{aligned}
p_{\text{slice}}:\;\;&\text{``You are a radiology expert analyzing this brain sMRI slice.} \\
&\text{Summarize your findings related to Alzheimer’s disease–related brain regions.''}
\end{aligned}
\]

In addition, we generate a volume-level description by presenting the set of selected slice images $\{\mathbf{I}_z\}_{z\in\mathcal{I}_k}$ jointly:
\[
\mathbf{c}_{\text{vol}} \;=\; f_{\text{VLM}}\big(\{\mathbf{I}_z\}_{z\in\mathcal{I}_k},\; p_{\text{vol}}\big),
\]
with the volume prompt
\[
\begin{aligned}
p_{\text{vol}}:\;\;&\text{``You are a radiology expert analyzing brain sMRI slices. Summarize the severity of brain region } \\
&\text{findings related to Alzheimer’s disease. Output one short paragraph.''}
\end{aligned}
\]

\paragraph{Text embedding.}
We embed the generated texts using a biomedical text encoder $f_{\text{txt}}(\cdot)$ (BiomedBERT) and $\ell_2$-normalize the features.
Let $\mathbf{t}_1,\dots,\mathbf{t}_{k+1}$ denote the ordered list of texts
\[
(\mathbf{t}_1,\dots,\mathbf{t}_k,\mathbf{t}_{k+1})
\;=\;
(\mathbf{c}_{z_1},\dots,\mathbf{c}_{z_k},\mathbf{c}_{\text{vol}}).
\]
The corresponding text embeddings are
\[
\mathbf{e}_i \;=\; 
\frac{f_{\text{txt}}(\mathbf{t}_i)}{\left\lVert f_{\text{txt}}(\mathbf{t}_i)\right\rVert_2}
\in\mathbb{R}^{d},
\qquad i\in[k+1],
\]
and we collect them into a matrix $\mathbf{E}\in\mathbb{R}^{(k+1)\times d}$ by stacking row-wise:
\[
\mathbf{E} \;=\; 
\begin{bmatrix}
\mathbf{e}_1^\top\\
\vdots\\
\mathbf{e}_{k+1}^\top
\end{bmatrix}.
\]
We save $\mathbf{E}$ together with the selected slice indices $\mathcal{I}_k$ for downstream training.

\section{Slice selection}
\label{appendix:selection}
Since sMRIs usually contain a large number of slices and the top and bottom ones often lack meaningful anatomical information, we adopt an intensity-based slice selection mechanism. 
Let $\mathcal{V}\in[0,1]^{Z\times H\times W}$ denote a normalized sMRI volume with $Z$ axial slices, and let $\mathcal{V}_z\in[0,1]^{H\times W}$ be the $z$-th slice. 
We first restrict attention to a central slice range by cropping along the $z$-axis. For a cropping ratio $\boldsymbol{\zeta}=(\zeta_0,\zeta_1)$ with $0\le \zeta_0<\zeta_1\le 1$, define
\[
z_0=\lfloor \zeta_0 Z\rfloor,\qquad z_1=\lfloor \zeta_1 Z\rfloor,\qquad \mathcal{Z}=\{z_0,\dots,z_1-1\}.
\]
For each $z\in\mathcal{Z}$, we compute a slice informativeness score
\[
s(z)\;=\;\frac{1}{3}\Big(H(z)+G(z)+V(z)\Big),
\]
where $H(z)$ is the entropy of the intensity histogram of $\mathcal{V}_z$, $G(z)$ is the mean gradient magnitude, and $V(z)$ is the intensity variance. Specifically, using $B$ histogram bins and letting $\mathbf{h}(z)\in\mathbb{R}^B$ be the normalized histogram with $\sum_{b=1}^B h_b(z)=1$,
\[
H(z)\;=\;-\sum_{b=1}^B h_b(z)\log\big(h_b(z)+\varepsilon\big),
\]
\[
G(z)\;=\;\frac{1}{HW}\sum_{i=1}^H\sum_{j=1}^W 
\sqrt{\big(\nabla_x \mathcal{V}_z(i,j)\big)^2+\big(\nabla_y \mathcal{V}_z(i,j)\big)^2},
\]
\[
V(z)\;=\;\frac{1}{HW}\sum_{i=1}^H\sum_{j=1}^W\Big(\mathcal{V}_z(i,j)-\mu_z\Big)^2,
\qquad 
\mu_z=\frac{1}{HW}\sum_{i=1}^H\sum_{j=1}^W \mathcal{V}_z(i,j).
\]
Finally, given $k\in[Z]$, we select the indices of the $k$ highest-scoring slices:
\[
\mathcal{I}_k \;=\;\operatorname{TopK}\big(\{s(z)\}_{z\in\mathcal{Z}}\big)\;\subseteq\;\mathcal{Z},
\]
and use the corresponding slice set $\{\mathcal{V}_z: z\in\mathcal{I}_k\}$ as the input to subsequent stages.

\section{Implementation}
\label{appendix:implementation}
\subsection{Implementation Details}

MVSC takes multiple preprocessed views of a 3D sMRI volume (\eg intensity-normalized images, brain masks, and tissue segmentations) as input, where a learnable residual augmentation based on the volume-wise mean is applied to the MRI features prior to MVSC processing.
Along with slice-level text and global text, slice-wise visual features are then extracted and stacked to form a unified representation.

\textbf{Encoding.}
For encoding, we decompose each slice into $16\times16$ patches and use a slice-level 2D CNN with kernel size and stride both set to 16 to extract non-overlapping patch features from each slice. The patch embeddings are combined with a channel-wise global volume summary via a weighted sum and serve as the input to subsequent modules.

\textbf{VoCE.}
The resulting patch embeddings, together with the global text, are processed by the VoCE module to model cross-patch and cross-slice interactions and produce a global volume-level feature.

\textbf{ASF.}
Slice-level fusion is then performed by the ASF module, where slice-level text is fused into the corresponding patch features. ASF aggregates multimodal patch features across slices to construct a visual surrogate representation, enabling both direct patch-level contributions and indirect cross-slice dependencies using the global feature.

\textbf{Decoding.}
The visual surrogate is reshaped and decoded into a three-channel feature using a symmetric slice-level 2D CNN with the same kernel size and stride as the encoder, mapping the patch embeddings back to the original spatial resolution. The decoded representation is then processed by a pretrained 2D vision foundation model for feature extraction.

\textbf{Classification.}
For downstream classification, we use a lightweight MLP head with a cross-entropy loss. The classifier consists linear layers with batch normalization, mapping the backbone-dependent feature dimension $N$ to the output classes ($N \rightarrow 512 \rightarrow 256 \rightarrow C$), where $C$ denotes the number of classes.

All trainable components are optimized using AdamW. Text embeddings are extracted using two NVIDIA H200 (141\,GB) GPUs, while all model training and inference are performed on a single NVIDIA V100 (32\,GB) GPU with a batch size of 8.

\subsection{Hyperparameter Settings}
\label{appendix:hyperparams}

MVSC is designed to adjust its key hyperparameters to accommodate datasets of different scales and complexities, enabling a balanced trade-off between model capacity and representation ability. In particular, the number of global features $K$ controls the capacity of the global volumetric representation and allows VoCE to capture multiple complementary aspects of volumetric structure. The feature dimension $d$ determines the representation capacity of each token, while the patch size controls the granularity at which local visual patterns are modeled.

Table~\ref{tab:mvsc_params} summarizes the hyperparameter configurations used for binary AD classification across datasets. Larger datasets such as ADNI adopt higher-capacity configurations, whereas smaller datasets such as AIBL and OASIS use lightweight settings to reduce overfitting and improve training stability.

\begin{table}[ht]
    \centering
    \small
    \caption{MVSC parameter configurations for AD classification across datasets.}
    \setlength{\tabcolsep}{6pt}
    \begin{tabular}{lcccc}
        \toprule
        Dataset & $K$ & $d$ & Patch size & Trainable Params \\
        \midrule
        ADNI  & 128 & 512 & 32 & 9.81M \\
        AIBL  & 32  & 128 & 16 & 0.70M \\
        OASIS & 32  & 128 & 16 & 0.70M \\
        \bottomrule
    \end{tabular}
    \label{tab:mvsc_params}
\end{table}


\end{document}